# Similarity Learning for Provably Accurate Sparse Linear Classification

Aurélien Bellet                                                                     AURELIEN.BELLET@UNIV-ST-ETIENNE.FR
Amaury Habrard                                                                    AMAURY.HABRARD@UNIV-ST-ETIENNE.FR
Marc Sebban                                                                          MARC.SEBBAN@UNIV-ST-ETIENNE.FR
Laboratoire Hubert Curien UMR CNRS 5516, Université Jean Monnet, 42000 Saint-Etienne, France

## Abstract

In recent years, the crucial importance of metrics in machine learning algorithms has led to an increasing interest for optimizing distance and similarity functions. Most of the state of the art focus on learning Mahalanobis distances (requiring to fulfill a constraint of positive semi-definiteness) for use in a *local* $k$-NN algorithm. However, no theoretical link is established between the learned metrics and their performance in classification. In this paper, we make use of the formal framework of $(\epsilon, \gamma, \tau)$-good similarities introduced by Balcan et al. to design an algorithm for learning a non PSD linear similarity optimized in a nonlinear feature space, which is then used to build a *global* linear classifier. We show that our approach has uniform stability and derive a generalization bound on the classification error. Experiments performed on various datasets confirm the effectiveness of our approach compared to state-of-the-art methods and provide evidence that (i) it is fast, (ii) robust to overfitting and (iii) produces very sparse classifiers.

## 1. Introduction

The notion of (dis)similarity plays an important role in many machine learning problems such as classification, clustering or ranking. For this reason, researchers have studied, in practical and formal ways, what it means for a pairwise similarity function to be "good". Since manually tuning such functions can be difficult and tedious for real-world problems, a lot of work has gone into automatically learning them from labeled data, leading to the emergence of *supervised similarity and metric learning*. Generally speaking, these approaches are based on the reasonable intuition that a good similarity function should assign a large (resp. small) score to pairs of points of the same class (resp. different class). Following this idea, they aim at finding the parameters (usually a matrix) of the function such that it satisfies best these *local pair-based constraints*. Among these methods, Mahalanobis distance learning (Schultz & Joachims, 2003; Shalev-Shwartz et al., 2004; Davis et al., 2007; Jain et al., 2008; Weinberger & Saul, 2009; Ying et al., 2009) has attracted a lot of interest, because it has a nice geometric interpretation: the goal is to learn a positive semi-definite (PSD) matrix which linearly projects the data into a new feature space where the standard Euclidean distance performs well. Some work has also gone into learning arbitrary similarity functions with no PSD constraint to make the problem easier to solve (Chechik et al., 2009; Qamar, 2010). The (dis)similarities learned with the above-mentioned methods are typically plugged in a $k$-Nearest Neighbor ($k$-NN) classifier (whose decision rule is based on a *local* neighborhood) and often lead to greater accuracy than the standard Euclidean distance, although no theoretical evidence supports this behavior. However, it seems unclear whether they can be successfully used in the context of *global* classifiers, such as linear separators.

Recently, Balcan et al. (2008) introduced the formal notion of $(\epsilon, \gamma, \tau)$-*good similarity function*, which does not require positive semi-definiteness and is less restrictive than local pair-based constraints. Indeed, it basically says that for most points, the *average* similarity scores to *some* points of the same class should be greater than to *some* points of different class. Assuming this property holds, generalization guarantees can be derived in terms of the error of a sparse linear classifier built from this similarity function.

In this paper, we use the notion of $(\epsilon, \gamma, \tau)$-goodness to design a new similarity learning algorithm. This novel approach, called SLLC (Similarity Learning for Linear Classification), has several advantages: (i) it is tailored to linear classifiers, (ii) theoretically well-founded, (iii) does not require positive semi-definiteness, and (iv) is in a sense less restrictive than pair-based settings. We formulate the problem of learning a good similarity function as an efficient convex quadratic program which optimizes a bilinear similarity. Furthermore, by using the Kernel Principal Component Analysis (KPCA) trick (Chatpatanasiri et al., 2010), we are able to kernelize our algorithm and thereby learn more powerful similarity functions and classifiers in the nonlinear feature space induced by a kernel. On the theoretical point of view, we show that our approach has uniform stability (Bousquet & Elisseeff, 2002), which leads to generalization guarantees in terms of the $(\epsilon, \gamma, \tau)$-goodness of the learned similarity. Lastly, we provide an experimental study on seven datasets of various domains and compare SLLC with two widely-used metric learning approaches. This study demonstrates the practical effectiveness of our method and shows that it is fast, robust to overfitting and induces very sparse classifiers, making it suitable for dealing with high-dimensional data.

The rest of the paper is organized as follows. Section 2 reviews some past work in similarity and metric learning and introduces the theory of $(\epsilon, \gamma, \tau)$-good similarities. Section 3 presents our approach, SLLC, and the KPCA trick used to kernelize it. Section 4 provides a theoretical analysis of SLLC, leading to the derivation of generalization guarantees. Finally, Section 5 features an experimental study on various datasets.

## 2. Notations and Related Work

We denote vectors by lower-case bold symbols ($\mathbf{x}$) and matrices by upper-case bold symbols ($\mathbf{A}$). We consider labeled points $\mathbf{z} = (\mathbf{x}, \ell)$ drawn from an unknown distribution $P$ over $\mathbb{R}^d \times \{-1, 1\}$. A similarity function is defined by $K : \mathbb{R}^d \times \mathbb{R}^d \to [-1, 1]$. We denote the $L_2$-norm by $\|\cdot\|_2$ and the Frobenius norm by $\|\cdot\|_\mathcal{F}$. Lastly, $[1-c]_+ = \max(0, 1-c)$ denotes the hinge loss.

### 2.1. Metric and Similarity Learning

Supervised metric and similarity learning aims at finding the parameters (usually a matrix) of a (dis)similarity function such that it best satisfies local constraints derived from the class labels. These constraints are typically *pair-based* ("examples $\mathbf{x}$ and $\mathbf{x}'$ should be similar/dissimilar") or *triplet-based* ("$\mathbf{x}$ should be more similar to $\mathbf{x}'$ than to $\mathbf{x}''$"). A great deal of work has focused on learning a (squared) Mahalanobis distance defined by $d^2_\mathbf{M}(\mathbf{x}, \mathbf{x}') = (\mathbf{x} - \mathbf{x}')^T \mathbf{M} (\mathbf{x} - \mathbf{x}')$ and parameterized by the PSD matrix $\mathbf{M} \in \mathbb{R}^{d \times d}$. A Mahalanobis distance implicitly corresponds to computing the Euclidean distance after some linear projection of the data. The PSD constraint ensures that this interpretation holds and makes $d_\mathbf{M}$ a (pseudo)metric, which enables $k$-NN speed-ups based on (for instance) the triangle inequality. The methods available in the literature mainly differ by their choices of objective/loss function and regularizer on $\mathbf{M}$. For instance, Schultz & Joachims (2003) require examples to be closer to similar examples than to dissimilar ones by a certain margin. Weinberger & Saul (2009) define an objective function related to the $k$-NN error on the training set. Davis et al. (2007) regularize using the LogDet divergence (for its automatic enforcement of PSD) while Ying et al. (2009) use the (2,1)-norm (which favors a low-rank $\mathbf{M}$). There also exist purely online methods (Shalev-Shwartz et al., 2004; Jain et al., 2008). The bottleneck of many of these approaches is to enforce the PSD constraint on $\mathbf{M}$, although some manage to reduce this computational burden by developing specific solvers. There has also been some interest in learning arbitrary similarity functions with no PSD requirement (Chechik et al., 2009; Qamar, 2010). All of the previously-mentioned learned similarities are used in the context of nearest-neighbors approaches (sometimes in clustering), which are based on local neighborhoods. In practice, they often outperform standard similarities, although no theoretical argument supports this behavior.

However, these *local* constraints do not seem appropriate to learn a similarity function for use in *global* classifiers, such a linear separators. The theory presented in the next section introduces a new, different notion of the goodness of a similarity function, and shows that such a good similarity achieves bounded error in linear classification. This opens the door to similarity learning for improving linear classifiers.

### 2.2. Learning with Good Similarity Functions

In recent work, Balcan et al. (2008) introduced a new theory of learning with *good similarity functions*, based on the following definition.

**Definition 1 (Balcan et al., 2008)** *A similarity function $K$ is an $(\epsilon, \gamma, \tau)$-good similarity function in hinge loss for a learning problem $P$ if there exists a (random) indicator function $R(\mathbf{x})$ defining a (probabilistic) set of "reasonable points" such that the* *following conditions hold:*

1. $\mathbb{E}_{(\mathbf{x},\ell)\sim P}[[1 - \ell g(\mathbf{x})/\gamma]_+] \leq \epsilon$,
   *where* $g(\mathbf{x}) = \mathbb{E}_{(\mathbf{x}',\ell')\sim P}[\ell' K(\mathbf{x},\mathbf{x}')|R(\mathbf{x}')]$,

2. $\Pr_{\mathbf{x}'}[R(\mathbf{x}')] \geq \tau$.

Thinking of this definition in terms of number of margin violations, we can interpret the first condition as "an $\epsilon$ proportion of examples $\mathbf{x}$ are on average $2\gamma$ more similar to random reasonable examples of the same class than to random reasonable examples of the opposite class" and the second condition as "at least a $\tau$ proportion of the examples should be reasonable". This definition is interesting in three respects. First, it does not impose positive semi-definiteness nor symmetry, which are requirements that may rule out the most natural similarity functions for some tasks. Second, it is based on an average over some points, which is less restrictive than pair or triplet-based settings. Third, satisfying Definition 1 is sufficient to learn well (Theorem 1).

**Theorem 1 (Balcan et al., 2008)** *Let $K$ be an $(\epsilon, \gamma, \tau)$-good similarity function in hinge loss for a learning problem $P$. For any $\epsilon_1 > 0$ and $0 \leq \delta \leq \gamma\epsilon_1/4$, let $S = \{\mathbf{x}'_1, \ldots, \mathbf{x}'_{d_{land}}\}$ be a (potentially unlabeled) sample of $d_{land} = \frac{2}{\tau}\left(\log(2/\delta) + 16\frac{\log(2/\delta)}{(\epsilon_1\gamma)^2}\right)$ landmarks drawn from $P$. Consider the mapping $\phi^S : \mathbb{R}^d \to \mathbb{R}^{d_{land}}$ defined as follows: $\phi^S_i(\mathbf{x}) = K(\mathbf{x}, \mathbf{x}'_i)$, $i \in \{1, \ldots, d_{land}\}$. Then, with probability at least $1 - \delta$ over the random sample $S$, the induced distribution $\phi^S(P)$ in $\mathbb{R}^{d_{land}}$ has a linear separator $\boldsymbol{\alpha}$ of error at most $\epsilon + \epsilon_1$ at margin $\gamma$.*

In other words, if we are given an $(\epsilon, \gamma, \tau)$-good similarity function for a learning problem $P$ and enough points (the "landmarks"), there exists a linear separator $\boldsymbol{\alpha}$ with error arbitrary close to $\epsilon$, which lies in the explicit "$\phi$-space" (the space of the similarity scores to the landmarks). As Balcan et al. mention, using $d_u$ (potentially unlabeled) landmark examples and $d_l$ labeled examples, we can estimate this separator $\boldsymbol{\alpha} \in \mathbb{R}^{d_u}$ by solving the following linear program:[1]

$$\min_{\boldsymbol{\alpha}} \sum_{i=1}^{d_l} \left[1 - \sum_{j=1}^{d_u} \alpha_j \ell_i K(\mathbf{x}_i, \mathbf{x}'_j)\right]_+ + \lambda\|\boldsymbol{\alpha}\|_1. \quad (1)$$

Note that Problem (1) is essentially an $L_1$-regularized linear SVM (Zhu et al., 2003) with an *empirical similarity map* (Balcan et al., 2008), and can be efficiently solved. The $L_1$-regularization induces sparsity (zero coordinates) in $\boldsymbol{\alpha}$, which reduces the number of training examples the classifier is based on, speeding up prediction. We can control the amount of sparsity by using the parameter $\lambda$ (the larger $\lambda$, the sparser $\boldsymbol{\alpha}$).

---

[1] The original formulation proposed by Balcan et al. (2008) was actually $L_1$-constrained. We turned it into an equivalent $L_1$-regularized one.

To sum up, the performance of the linear classifier theoretically depends on how well the similarity function satisfies Definition 1. However, for some problems, standard similarity functions may satisfy the definition poorly, leading to weak guarantees. To deal with this limitation, Kar & Jain (2011) propose to automatically adapt the goodness criterion to the problem at hand. In this paper, we take a different approach: we see Definition 1 as a novel, theoretically well-founded objective function for similarity learning.

## 3. Learning Good Similarity Functions for Linear Classification

We consider $K_\mathbf{A}(\mathbf{x}, \mathbf{x}') = \mathbf{x}^T \mathbf{A} \mathbf{x}'$, a *bilinear similarity* parameterized by the matrix $\mathbf{A} \in \mathbb{R}^{d \times d}$, which is not constrained to be PSD nor symmetric. This form of similarity function was successfully used in the context of large-scale online image similarity learning (Chechik et al., 2009). $K_\mathbf{A}$ has the advantage of being efficiently computable when the inputs $\mathbf{x}$ and $\mathbf{x}'$ are sparse vectors. In order to satisfy the condition $K_\mathbf{A} \in [-1, 1]$, we assume that inputs are normalized such that $\|\mathbf{x}\|_2 \leq 1$, and we require $\|\mathbf{A}\|_\mathcal{F} \leq 1$.

### 3.1. Similarity Learning Formulation

Our goal is to optimize the $(\epsilon, \gamma, \tau)$-goodness of $K_\mathbf{A}$ on a finite-size sample. To this end, we are given a training sample of $N_T$ labeled points $T = \{\mathbf{z}_i = (\mathbf{x}_i, \ell_i)\}_{i=1}^{N_T}$ and a sample of $N_R$ labeled reasonable points $R = \{\mathbf{z}_k = (\mathbf{x}_k, \ell_k)\}_{k=1}^{N_R}$. In practice, $R$ is a subset of $T$ with $N_R = \hat{\tau} N_T$ ($\hat{\tau} \in ]0,1]$). In the lack of background knowledge, it can be drawn randomly or according to some criterion (e.g., diversity (Kar & Jain, 2011)). Given $R$ and a margin $\gamma$, let also $V(\mathbf{A}, \mathbf{z}_i, R) = [1 - \ell_i \frac{1}{\gamma N_R} \sum_{k=1}^{N_R} \ell_k K_\mathbf{A}(\mathbf{x}_i, \mathbf{x}_k)]_+$ denote the empirical goodness of $K_\mathbf{A}$ with respect to a single training point $\mathbf{z}_i$, and $\epsilon_T = \frac{1}{N_T} \sum_{i=1}^{N_T} V(\mathbf{A}, \mathbf{z}_i, R)$ the empirical goodness over $T$.

Now, we want to learn the matrix $\mathbf{A}$ that minimizes $\epsilon_T$. This can be done by solving the following regularized problem, referred to as SLLC (Similarity Learning for Linear Classification):

$$\min_{\mathbf{A} \in \mathbb{R}^{d \times d}} \epsilon_T + \beta \|\mathbf{A}\|_\mathcal{F}^2$$

where $\beta$ is a regularization parameter. Note that an equivalent constrained formulation can be obtained by rewriting the sum of $N_T$ hinge losses in the objective function as $N_T$ margin constraints and introducing $N_T$ slack variables in the objective.

SLLC is radically different from classic metric and similarity learning algorithms, which are based on pair or triplet-based constraints. It learns a global similarity rather than a local one, since $R$ is the same for each training example. Moreover, the constraints are easier to satisfy since they are defined over an *average* of similarity scores to the points in $R$ instead of over a single pair or triplet. This means that one can fulfill a constraint without satisfying the margin for each point in $R$ individually. SLLC has also a number of desirable properties: (i) This is a convex quadratic program, which can be solved efficiently using standard solvers. No costly semi-definite programming is required, as opposed to many Mahalanobis distance learning methods. (ii) In the constrained formulation, there is only one constraint per training example (instead of one for each pair or triplet), i.e., a total of $N_T$ constraints and $N_T + d^2$ variables. (iii) The size of $R$ does not affect the complexity of the constraints. (iv) If $\mathbf{x_i}$ is sparse, then the associated constraint is sparse as well (some variables of the problem do not appear).

### 3.2. Kernelization of SLLC

The framework presented in the previous section is theoretically well-founded with respect to Balcan et al.'s theory and has some generalization guarantees, as we will see in the next section. Moreover, it has the advantage of being very simple: we learn a global linear similarity and use it to build a global linear classifier. In order to learn more powerful similarities (and therefore classifiers), we propose to kernelize the approach by learning them in the nonlinear feature space induced by a kernel. Kernelization allows linear classifiers such as Support Vector Machines or some Mahalanobis distance learning algorithms (e.g., Shalev-Shwartz et al., 2004; Davis et al., 2007) to learn nonlinear decision boundaries or transformations. However, kernelizing a metric learning algorithm is not trivial: a new formulation of the problem has to be derived, where interface to the data is limited to inner products, and sometimes a different implementation is necessary. Moreover, when kernelization is possible, one must learn a $N_T \times N_T$ matrix. As $N_T$ gets large, the problem becomes intractable unless dimensionality reduction is applied.

For these reasons, we instead use the KPCA trick, recently proposed by Chatpatanasiri et al. (2010). It provides a straightforward way to kernelize a metric learning algorithm while performing dimensionality reduction at no additional cost. The idea is to use Kernel Principal Component Analysis (Schölkopf et al., 1998) to project the data into a new space using a nonlinear kernel function, and to keep only a chosen number of dimensions (those that capture best the overall variance of the data). The data are then projected into this new feature space, and the (unchanged) metric learning algorithm can be used to learn a metric in that space. Chatpatanasiri et al. (2010) showed that the KPCA trick is theoretically sound for unconstrained metric and similarity learning algorithms (they proved representer theorems), which includes SLLC. Throughout the rest of this paper, we will only consider the kernelized version of SLLC.

Generally speaking, kernelizing a metric learning algorithm may cause or increase overfitting, especially when data are scarce and/or high-dimensional. However, since our entire framework is linear and global, we expect our method to be quite robust to this effect. This will be doubly confirmed in the rest of this paper: experimentally in Section 5, but also theoretically with the derivation in the following section of generalization guarantees independent from the size of the projection space.

## 4. Theoretical Analysis

In this section, we present a theoretical analysis of our approach. Our main result is the derivation of a generalization bound (Theorem 3) guaranteeing the consistency of SLLC and thus the $(\epsilon, \gamma, \tau)$-goodness for the considered task. In our framework, the similarity is optimized according to a set $R$ of reasonable points coming from the training sample. Therefore, these reasonable points may not follow the distribution from which the training sample has been generated. To cope with this situation, we propose to derive a generalization bound according to the framework of *uniform stability* (Bousquet & Elisseeff, 2002), which does not assume an i.i.d. draw at the pair level.

### 4.1. Uniform Stability

Roughly speaking, an algorithm is *stable* if its output does not change significantly under a small modification of the training sample. This variation is required to be bounded in $\mathcal{O}(1/N_T)$ in terms of infinite norm.

**Definition 2 (Bousquet & Elisseeff, 2002)** *A learning algorithm has a uniform stability in $\frac{\kappa}{N_T}$ w.r.t. a loss function $\mathcal{L}$, with $\kappa$ a positive constant, if*

$$\forall T, \forall i, 1 \leq i \leq N_T, \sup_{\mathbf{z}} |\mathcal{L}(\mathbf{M}_T, \mathbf{z}) - \mathcal{L}(\mathbf{M}_{T^i}, \mathbf{z})| \leq \frac{\kappa}{N_T},$$

*where $\mathbf{M}_T$ is the model learned from the sample $T$,*

Similarity Learning for Provably Accurate Sparse Linear Classification$\mathbf{M}_{T^i}$ the model learned from the sample $T^i$. $T^i$ is obtained from $T$ by replacing the $i^{th}$ example $\mathbf{z}_i \in T$ by another example $\mathbf{z}'_i$ independent from $T$ and drawn from $P$. $\mathcal{L}(\mathbf{M}, \mathbf{z})$ is the loss for an example $\mathbf{z}$.

In this definition, $T^i$ characterizes the notion of small modification of the training sample. When this definition is fulfilled, Bousquet & Elisseeff (2002) have shown that the following generalization bound holds.

**Theorem 2 (Bousquet & Elisseeff, 2002)** *Let $\delta > 0$ and $N_T > 1$. For any algorithm with uniform stability $\kappa/N_T$ using a loss function bounded by 1, with probability $1 - \delta$ over the random draw of $T$:*

$$L(\mathbf{M}_T) < \hat{L}_T(\mathbf{M}_T) + \frac{\kappa}{N_T} + (2\kappa + 1)\sqrt{\frac{\ln 1/\delta}{2N_T}},$$

*where $L(\mathbf{M}_T)$ is the expected loss and $\hat{L}_T(\mathbf{M}_T)$ its empirical estimate over $T$.*

### 4.2. Generalization Bound

For convenience, given a bilinear model $K_\mathbf{A}$, we denote by $\mathbf{A}_R$ both the similarity defined by the matrix $\mathbf{A}$ and its associated set of reasonable points $R$ (when it is clear from the context we may omit the subscript $R$). Given a similarity $\mathbf{A}_R$, $V(\mathbf{A}_R, \mathbf{z}, R)$ plays the role of the loss function over one example $\mathbf{z}$. The loss over the sample $T$ is defined as $\epsilon_T(\mathbf{A}_R) = \frac{1}{N_T}\sum_{i=1}^{N_T} V(\mathbf{A}_R, \mathbf{z}_i, R)$, and corresponds to the empirical goodness. Lastly, the expected loss over the true distribution is given by $\epsilon(\mathbf{A}_R) = \mathbb{E}_{\mathbf{z}=(\mathbf{x},l)\sim P} V(\mathbf{A}_R, \mathbf{z}, R)$, and corresponds to the goodness in generalization. When it is clear from the context we may simply use $\epsilon_T$ and $\epsilon$.

In our case, to prove the uniform stability property we need to show that

$$\forall T, \forall i, \sup_\mathbf{z} |V(\mathbf{A}, \mathbf{z}, R) - V(\mathbf{A}^i, \mathbf{z}, R^i)| \leq \frac{\kappa}{N_T},$$

where $\mathbf{A}$ is learned from $T$, $R \subseteq T$, $\mathbf{A}^i$ is the matrix learned from $T^i$ and $R^i \subseteq T^i$ is the set of reasonable points associated to $T^i$. Note that $R$ and $R^i$ are of equal size and can differ in at most one example, depending on whether $\mathbf{z}_i$ or $\mathbf{z}'_i$ belong to their corresponding set of reasonable points. For the sake of simplicity, we consider $V$ bounded by 1 (which can be easily obtained by dividing it by the constant $1 + \frac{1}{\gamma}$). To show this property, we need the following results.

**Lemma 1** *For any labeled examples $\mathbf{z} = (\mathbf{x}, \ell)$, $\mathbf{z}' = (\mathbf{x}', \ell')$ and any models $\mathbf{A}_R$, $\mathbf{A}'_{R'}$, the following holds:*

*P1:* $|K_\mathbf{A}(\mathbf{x}, \mathbf{x}')| \leq 1$,

*P2:* $|K_\mathbf{A}(\mathbf{x}, \mathbf{x}') - K_{\mathbf{A}'}(\mathbf{x}, \mathbf{x}')| \leq \|\mathbf{A} - \mathbf{A}'\|_\mathcal{F}$,

*P3:* $|V(\mathbf{A}, \mathbf{z}, R) - V(\mathbf{A}', \mathbf{z}, R')| \leq 1 |\frac{\sum_{k=1}^{N_R} \ell_k K_\mathbf{A}(\mathbf{x}, \mathbf{x}_k)}{\gamma N_R} - \frac{\sum_{j=1}^{N_{R'}} \ell'_k K_{\mathbf{A}'}(\mathbf{x}, \mathbf{x}'_k)}{\gamma N_{R'}}|$ *(1-admissibility property of $V$).*

**Proof** P1 comes from $|K_\mathbf{A}(\mathbf{x}, \mathbf{x}')| \leq \|\mathbf{x}\|_2 \|\mathbf{A}\|_\mathcal{F} \|\mathbf{x}'\|_2$, the normalization on examples ($\|\mathbf{x}\|_2 \leq 1$) and the requirement on matrices ($\|\mathbf{A}\|_\mathcal{F} \leq 1$).

For P2, we observe that $|K_\mathbf{A}(\mathbf{x}, \mathbf{x}') - K_{\mathbf{A}'}(\mathbf{x}, \mathbf{x}')| = |K_{\mathbf{A}-\mathbf{A}'}(\mathbf{x}, \mathbf{x}')|$, and we use the normalization $\|\mathbf{x}\|_2 \leq 1$.

P3 follows directly from $|\ell| = 1$ and the 1-lipschitz property of the hinge loss: $|[X]_+ - [Y]_+| \leq |X - Y|$. □

Let $F_T = \epsilon_T(\mathbf{A}) + \beta \|\mathbf{A}\|_\mathcal{F}^2$ be the objective function of SLLC w.r.t. a sample $T$ and a set of reasonable points $R \subseteq T$. The following lemma bounds the deviation between $\mathbf{A}$ and $\mathbf{A}^i$.

**Lemma 2** *For any models $\mathbf{A}$ and $\mathbf{A}^i$ that are minimizers of $F_T$ and $F_{T^i}$ respectively, we have:*

$$\|\mathbf{A} - \mathbf{A}^i\|_\mathcal{F} \leq \frac{1}{\beta N_T \gamma}.$$

**Proof** We follow closely the proof of Lemma 20 of (Bousquet & Elisseeff, 2002) and omit some details due to the limitation of space. Let $\Delta \mathbf{A} = \mathbf{A}^i - \mathbf{A}$ and $0 \leq t \leq 1$, $M_1 = \|\mathbf{A}\|_\mathcal{F}^2 - \|\mathbf{A} + t\Delta\mathbf{A}\|_\mathcal{F}^2 + \|\mathbf{A}^i\|_\mathcal{F}^2 - \|\mathbf{A}^i - t\Delta\mathbf{A}\|_\mathcal{F}^2$ and $M_2 = \frac{1}{\beta N_T}(\epsilon_T(\mathbf{A}_R) - \epsilon_T((\mathbf{A}+t\Delta\mathbf{A})_R) + \epsilon_{T^i}((\mathbf{A}+t\Delta\mathbf{A})_R) - \epsilon_{T^i}(\mathbf{A}_R))$. Using the fact that $F_T$ and $F_{T^i}$ are convex functions, that $\mathbf{A}$ and $\mathbf{A}^i$ are their respective minimizers and property P3, we have $M_1 \leq M_2$. Fixing $t = 1/2$, we obtain $M_1 = \|\mathbf{A} - \mathbf{A}^i\|_\mathcal{F}^2$, and using property P3 and the normalization $\|\mathbf{x}\|_2 \leq 1$, we get:

$$M_2 \leq \frac{1}{\beta N_T \gamma}(\|\frac{1}{2}\Delta\mathbf{A}\|_\mathcal{F} + \|-\frac{1}{2}\Delta\mathbf{A}\|_\mathcal{F}) = \frac{\|\mathbf{A} - \mathbf{A}^i\|_\mathcal{F}}{\beta N_T \gamma}.$$

This leads to the inequality $\|\mathbf{A} - \mathbf{A}^i\|_\mathcal{F}^2 \leq \frac{\|\mathbf{A}-\mathbf{A}^i\|_\mathcal{F}}{\beta N_T \gamma}$ from which Lemma 2 is directly derived. □

We now have all the material needed to prove the stability property of our algorithm.

**Lemma 3** *Let $N_T$ and $N_R$ be the number of training examples and reasonable points respectively, $N_R = \hat{\tau} N_T$ with $\hat{\tau} \in ]0, 1]$. SLLC has a uniform stability in $\frac{\kappa}{N_T}$ with $\kappa = \frac{1}{\gamma}(\frac{1}{\beta\gamma} + \frac{2}{\hat{\tau}}) = \frac{\hat{\tau}+2\beta\gamma}{\hat{\tau}\beta\gamma^2}$, where $\beta$ is the regularization parameter and $\gamma$ the margin.*

**Proof** For any sample $T$ of size $N_T$, any $1 \leq i \leq N_T$, any labeled examples $\mathbf{z} = (\mathbf{x}, \ell)$ and $\mathbf{z}'_i = (\mathbf{x}_i, \ell'_i) \sim P$:

$$|V(\mathbf{A}, \mathbf{z}, R) - V(\mathbf{A}^i, \mathbf{z}, R^i)|$$
$$\leq \left|\frac{1}{\gamma N_R}\sum_{k=1}^{N_R}\ell_k K_\mathbf{A}(\mathbf{x}, \mathbf{x}_k) - \frac{1}{\gamma N_{R^i}}\sum_{k=1}^{N_{R^i}}\ell_k K_{\mathbf{A}^i}(\mathbf{x}, \mathbf{x}_k)\right|$$

$$\begin{aligned}
&= \left| \frac{1}{\gamma N_R} \left( \left( \sum_{k=1, k \neq i}^{N_R} (\ell_k K_{\mathbf{A}}(\mathbf{x}, \mathbf{x}_k) - K_{\mathbf{A}^i}(\mathbf{x}, \mathbf{x}_k)) \right) + \right. \right. \\
&\qquad \left. \left. \ell_i K_{\mathbf{A}}(\mathbf{x}, \mathbf{x}_i) - \ell'_i K_{\mathbf{A}^i}(\mathbf{x}, \mathbf{x}'_i) \right) \right| \\
&\leq \frac{1}{\gamma N_R} \left( \left( \sum_{k=1, k \neq i}^{N_R} (|\ell_k| \|\mathbf{A} - \mathbf{A}^i\|_{\mathcal{F}}) \right) + \right. \\
&\qquad \left. |\ell_i K_{\mathbf{A}^i}(\mathbf{x}, \mathbf{x}_i)| + |\ell'_i K_{\mathbf{A}}(\mathbf{x}, \mathbf{x}'_i)| \right) \\
&\leq \frac{1}{\gamma N_R} \left( \frac{N_R - 1}{\beta N_T \gamma} + 2 \right) \leq \frac{1}{\gamma N_R} \left( \frac{N_R}{\beta N_T \gamma} + 2 \right).
\end{aligned}$$

The first inequality follows from $P3$. The second comes from the fact that $R$ and $R^i$ differ in at most one element, corresponding to the example $\mathbf{z}_i$ in $R$ and the example $\mathbf{z}'_i$ replacing $\mathbf{z}_i$ in $R^i$. The last inequalities are obtained by the use of the triangle inequality, $P1$, $P2$, Lemma 2, and the fact that the labels belong to $\{-1, 1\}$. Since $N_R = \hat{\tau} N_T$, we get $|V(\mathbf{A}, \mathbf{z}, R) - V(\mathbf{A}^i, \mathbf{z}, R^i)| \leq \frac{1}{\gamma N_T}(\frac{1}{\beta\gamma} + \frac{2}{\hat{\tau}})$. □

Applying Thm 2 with Lemma 2 gives our main result.

**Theorem 3** *Let $\gamma > 0$, $\delta > 0$ and $N_T > 1$. With probability at least $1 - \delta$, for any model $\mathbf{A}_R$ learned with SLLC, we have:*

$$\epsilon \leq \epsilon_T + \frac{1}{N_T} \left( \frac{\hat{\tau} + 2\beta\gamma}{\hat{\tau}\beta\gamma^2} \right) + \left( \frac{2(\hat{\tau} + 2\beta\gamma)}{\hat{\tau}\beta\gamma^2} + 1 \right) \sqrt{\frac{\ln 1/\delta}{2 N_T}}.$$

Thm 3 highlights three important properties of SLLC. First, it converges in $\mathcal{O}(1/\sqrt{N_T})$, which is a standard convergence rate for uniform stability. Second, it is independent from the dimensionality of the data. This is due to the fact that $\|\mathbf{A}\|_{\mathcal{F}}$ is bounded by a constant. Third, Thm 3 bounds the goodness in generalization of the learned similarity function. By minimizing $\epsilon_T$ with SLLC, we minimize $\epsilon$ and thus the error of the resulting linear classifier, as stated by Thm 1.

## 5. Experiments

We propose a comparative study of our method and two widely-used Mahalanobis distance learning algorithms: Large Margin Nearest Neighbor (LMNN) from Weinberger & Saul (2009) and Information-Theoretic Metric Learning (ITML) from Davis et al. (2007).[2] Roughly speaking, LMNN optimizes the $k$-NN error on the training set (with a safety margin) whereas ITML aims at best satisfying pair-based constraints while minimizing the LogDet divergence between the learned matrix $\mathbf{M}$ and the identity matrix. We conduct this experimental study on seven classic binary datasets of varying domain, size and difficulty, mostly taken from the UCI Machine Learning Repository. Their properties are summarized in Table 1. Some of them, such as Breast, Ionosphere or Pima, have been extensively used to evaluate metric learning methods.

### 5.1. Experimental Setup

We compare the following methods: (i) the cosine similarity $K_I$ in KPCA space, as a baseline, (ii) SLLC, (iii) LMNN in the original space, (iv) LMNN in KPCA space, (v) ITML in the original space, and (vi) ITML in KPCA space.[3] All attributes are scaled to $[-1/d; 1/d]$ to ensure $\|\mathbf{x}\|_2 \leq 1$. To generate a new feature space using KPCA, we use the Gaussian kernel with parameter $\sigma$ equal to the mean of all pairwise training data Euclidean distances (a standard heuristic, used for instance by Kar & Jain (2011)). Ideally, we would like to project the data to the feature space of maximum size (equal to the number of training examples), but to keep the computations tractable we only retain three times the number of features of the original data (four times for the low-dimensional datasets), as shown in Table 1.[4] On Cod-RNA, KPCA was run on a randomly drawn subsample of 10% of the training data. Unless predefined training and test sets are available (as for Splice, Svmguide1 and Cod-RNA), we randomly generate 70/30 splits of the data, and average the results over 100 runs. Training sets are further partitioned 70/30 for validation purposes. We tune the following parameters by cross-validation: $\beta, \gamma \in \{10^{-7}, \ldots, 10^{-2}\}$ for SLLC, $\lambda_{ITML} \in \{10^{-4}, \ldots, 10^4\}$ for ITML, and $\lambda \in \{10^{-3}, \ldots, 10^2\}$ for learning the linear classifiers, choosing the value offering the best accuracy. We choose $R$ to be the entire training set, i.e., $\hat{\tau} = 1$ (interestingly, cross-validation of $\hat{\tau}$ did not improve the results significantly). We take $k = 3$ and $\mu = 0.5$ for LMNN, as done in (Weinberger & Saul, 2009). For ITML, we generate $N_T$ random constraints for a fair comparison with SLLC.

### 5.2. Results

**Classification performance** We report the results obtained with the sparse linear classifier of Problem (1) suggested by Balcan et al. (2008) but also those obtained with 3-NN since LMNN and ITML are designed

---
[2] We used the code provided by the authors.

[3] $K_I$, LMNN and ITML are normalized to ensure their values belong to $[-1, 1]$.

[4] Note that the amount of variance captured thereby was greater than 90% for all datasets.

Similarity Learning for Provably Accurate Sparse Linear Classification

Table 1. Properties of the seven datasets used in the experimental study.

|  | Breast | Iono. | Rings | Pima | Splice | Svmguide1 | Cod-RNA |
|---|---|---|---|---|---|---|---|
| # training examples | 488 | 245 | 700 | 537 | 1,000 | 3,089 | 59,535 |
| # test examples | 211 | 106 | 300 | 231 | 2,175 | 4,000 | 271,617 |
| # dimensions | 9 | 34 | 2 | 8 | 60 | 4 | 8 |
| # dim. after KPCA | 27 | 102 | 8 | 24 | 180 | 16 | 24 |
| # runs | 100 | 100 | 100 | 100 | 1 | 1 | 1 |

Table 2. Average accuracy (normal type) and sparsity (italic type) of the linear classifiers built from the studied similarity functions. For each dataset, bold font indicates the most accurate method (sparsity is used to break the ties).

|  | Breast | Iono. | Rings | Pima | Splice | Svmguide1 | Cod-RNA |
|---|---|---|---|---|---|---|---|
| $K_I$ | 96.57 *20.39* | 89.81 *52.93* | 100.00 *18.20* | 75.62 *25.93* | 83.86 *362* | **96.95** *64* | **95.91** *557* |
| SLLC | **96.90** *1.00* | **93.25** *1.00* | **100.00** *1.00* | **75.94** *1.00* | **87.36** *1* | 96.55 *8* | 94.08 *1* |
| LMNN | 96.81 *9.98* | 90.21 *13.30* | 100.00 *18.04* | 75.15 *69.71* | 85.61 *315* | 95.80 *157* | 88.40 *61* |
| LMNN KPCA | 96.01 *8.46* | 86.12 *9.96* | 100.00 *8.73* | 74.92 *22.20* | 86.85 *156* | 96.53 *82* | 95.15 *591* |
| ITML | 96.80 *9.79* | 92.09 *9.51* | 100.00 *17.85* | 75.25 *56.22* | 81.47 *377* | 96.70 *49* | 95.06 *164* |
| ITML KPCA | 96.23 *17.17* | 93.05 *18.01* | 100.00 *15.21* | 75.25 *16.40* | 85.29 *287* | 96.55 *89* | 95.14 *206* |

for $k$-NN use.[5] In linear classification (Table 2), SLLC achieves the highest accuracy on 5 out of 7 datasets and competitive performance on the remaining 2. At the same time, on all datasets, SLLC leads to extremely sparse classifiers. The sparsity of the classifier corresponds to the number of training examples that are involved in classifying a new example. Therefore, SLLC leads to much simpler and yet often more accurate classifiers than those built from other similarities. Furthermore, sparsity allows faster predictions, especially when data are plentiful and/or high-dimensional (e.g., Cod-RNA or Splice). Often enough, the learned linear classifier has sparsity 1, which means that classifying a new example boils down to computing its similarity score to a single training example and compare the value with a threshold. Note that we tried large values of $\lambda$ to obtain sparser classifiers from $K_I$, LMNN and ITML, but this yielded dramatic drops in accuracy. The extreme sparsity brought by SLLC comes from the fact that the constraints are based on an average of similarity scores over the same set of points for all training examples. Surprisingly, in 3-NN classification (Table 3), SLLC achieves the best results on 4 datasets. It is, however, outperformed by LMNN or ITML on the 3 biggest problems. For most tasks, the accuracy obtained in linear classification is better or similar to that obtained with 3-NN (highlighting the fact that similarity learning for linear classification is of interest) while prediction is many orders of magnitude faster due to the sparsity of the linear separators.

---
[5]When necessary, we take the opposite value of a similarity to get a measure of distance, and vice versa.

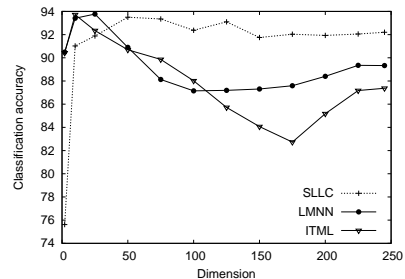

Figure 1. Accuracy of the methods with respect to the dimensionality of the KPCA space on Ionosphere.

Also note that a good similarity for $k$-NN classification can achieve poor results in linear classification (LMNN on Cod-RNA), and vice versa (SLLC on Svmguide1).

**Robustness to overfitting** The fact that SLLC performs well on small datasets is partly due to its robustness to overfitting, highlighted in Figure 1. As expected, LMNN and ITML, which are optimized locally and plugged in a local nonlinear classifier, tend to overfit as the dimensionality grows. On the other hand, SLLC suffers from very limited overfitting due to its global and linear setting.

**Time comparison** Note that SLLC is solved using the standard convex programming package MOSEK while LMNN and ITML have their own specific and sophisticated solver. Despite this fact, SLLC is several orders of magnitude faster than LMNN (see Table 4) because its number of constraints is much smaller.

Similarity Learning for Provably Accurate Sparse Linear Classification

Table 3. Average accuracy of 3-NN classifiers using the studied similarity functions.

|  | Breast | Iono. | Rings | Pima | Splice | Svmguide1 | Cod-RNA |
|---|---|---|---|---|---|---|---|
| $K_I$ | 96.71 | 83.57 | **100.00** | 72.78 | 77.52 | 93.93 | 90.07 |
| SLLC | **96.90** | **93.25** | **100.00** | **75.94** | 87.36 | 93.82 | 94.08 |
| LMNN | 96.46 | 88.68 | **100.00** | 72.84 | 83.49 | 96.23 | 94.98 |
| LMNN KPCA | 96.23 | 87.13 | **100.00** | 73.50 | **87.59** | 95.85 | 94.43 |
| ITML | 92.67 | 88.29 | **100.00** | 72.07 | 77.43 | 95.97 | **95.42** |
| ITML KPCA | 96.38 | 87.56 | **100.00** | 72.80 | 84.41 | **96.80** | 95.32 |

Table 4. Average time per run (in seconds) required for learning the similarity.

|  | Breast | Iono. | Rings | Pima | Splice | Svmguide1 | Cod-RNA |
|---|---|---|---|---|---|---|---|
| SLLC | 4.76 | 5.36 | 0.05 | 4.01 | 158.38 | 185.53 | 2471.25 |
| LMNN | 25.99 | 16.27 | 37.95 | 32.14 | 309.36 | 331.28 | 10418.73 |
| LMNN KPCA | 41.06 | 34.57 | 84.86 | 48.28 | 1122.60 | 369.31 | 24296.41 |
| ITML | 2.09 | 3.09 | 0.19 | 2.96 | 3.41 | 0.83 | 5.98 |
| ITML KPCA | 1.68 | 5.77 | 0.20 | 2.74 | 56.14 | 5.30 | 25.25 |

However, it remains slower than ITML.

## 6. Conclusion

In this paper, we presented SLLC, a novel approach to the problem of similarity learning by making use of both Balcan et al.'s theory of $(\epsilon, \gamma, \tau)$-good similarity functions and the KPCA trick. We derived a generalization bound based on the notion of uniform stability that is independent from the size of the input space, and thus from the number of dimensions selected by KPCA. It guarantees the goodness in generalization of the learned similarity, and therefore the accuracy of the resulting linear classifier for the considered task. We experimentally demonstrated the effectiveness of SLLC and also showed that the learned similarities induce extremely sparse classifiers. Combined with the independence from dimensionality and the robustness to overfitting, it makes the approach very efficient and suitable for high-dimensional data. Future work could include a "full" kernelization of SLLC (i.e., express the problem solely in terms of inner products), studying the influence of other regularizers on **M** (for instance, using the nuclear norm to learn low-rank matrices), developing a specific solver to match ITML's speed and the derivation of an online algorithm.

## Acknowledgments

We would like to acknowledge support from the ANR LAMPADA 09-EMER-007-02 project and the PASCAL 2 Network of Excellence.